\documentclass[10pt,twocolumn,letterpaper]{article}

\usepackage[pagenumbers]{cvpr}      

\usepackage{graphicx}
\usepackage{amsmath}
\usepackage{amssymb}
\usepackage{booktabs}

\usepackage{multirow}
\usepackage{amssymb}
\usepackage{pifont}

\usepackage[table]{xcolor}

\usepackage[pagebackref,breaklinks,colorlinks]{hyperref}

\usepackage[capitalize]{cleveref}
\crefname{section}{Sec.}{Secs.}
\Crefname{section}{Section}{Sections}
\Crefname{table}{Table}{Tables}
\crefname{table}{Tab.}{Tabs.}

\def\cvprPaperID{20} 
\def\confName{CVPR}
\def\confYear{2023}

\begin{document}

\title{An Extended Study of Human-like Behavior under Adversarial Training}
\author{
Paul Gavrikov$^{1}$\thanks{Funded by the Ministry for Science, Research and Arts, Baden-Wuerttemberg, Grant 32-7545.20/45/1 (Q-AMeLiA).} \qquad Janis Keuper$^{1,2}$\footnotemark[1] \qquad Margret Keuper$^{3,4}$\\
$^{1}$IMLA, Offenburg University, $^{2}$Fraunhofer ITWM, $^{3}$University of Siegen\\
$^{4}$Max Planck Institute for Informatics, Saarland Informatics Campus\\
{\tt\small \{paul.gavrikov,janis.keuper\}@hs-offenburg.de,  
keuper@mpi-inf.mpg.de}
}
\maketitle

\begin{abstract} 
Neural networks have a number of shortcomings. Amongst the severest ones is the sensitivity to distribution shifts which allows models to be easily fooled into wrong predictions by small perturbations to inputs that are often imperceivable to humans and do not have to carry semantic meaning. Adversarial training poses a partial solution to address this issue by training models on worst-case perturbations. Yet, recent work has also pointed out that the reasoning in neural networks is different from humans. Humans identify objects by shape, while neural nets mainly employ texture cues. Exemplarily, a model trained on photographs will likely fail to generalize to datasets containing sketches. Interestingly, it was also shown that adversarial training seems to favorably increase the shift toward shape bias. In this work, we revisit this observation and provide an extensive analysis of this effect on various architectures, the common $\ell_2$- and $\ell_\infty$-training, and Transformer-based models. Further, we provide a possible explanation for this phenomenon from a frequency perspective.
\end{abstract}

\section{Introduction}
\label{sec:intro}
ImageNet~\cite{imagenet} trained convolutional neural networks (CNNs) have been shown to predominantly classify images by the observed texture, whereas, humans rather tend to consider global object shapes as the predominant cues~\cite{geirhos2018imagenettrained}. In this context, \textit{Geirhos} \etal provided an initial analysis of robust models and provided initial evidence, that the initial texture bias in CNNs is shifted towards shape-based decisions under adversarial training (AT) \cite{Geirhos2021_modelsvshumasn}. However,  the authors have limited their analysis to a ResNet-50 trained on ImageNet using AT against an $\ell_2$-bound adversary. To allow for a more conclusive evaluation, we expand their analysis to the more common $\ell_\infty$-setting for AT and analyze additional CNNs like ResNet-18 and Wide-ResNet-50-2, as well as Transformers (XCiT-S/M/L12), which are known to behave differently from CNNs regarding their inductive bias. In our study, we evaluate models trained on clean data as well as under AT with different norms and budgets with respect to their generalization ability to out-of-domain (OOD) data \cite{Geirhos2021_modelsvshumasn,Wang2019Learning}, with special emphasis on the shape-texture \textit{cue-conflict}, that has been used as a measure of the misalignment between human and neural network based image classification. In this context, we provide an extensive evaluation and discussion of the different behavior of CNN and Transformer models. 

Further, we analyze the generalization ability of adversarially-trained networks from a frequency perspective. Specifically, we investigate the frequency spectra of different OOD image categories and provide possible explanations for the following two questions: (i) Why does adversarial training lead to an accuracy decay on some OOD datasets? and (ii) Why is the \textit{cue-conflict} between shape and texture affected by AT?

We summarize our key findings as follows:
\begin{itemize}
    \item Training against $\ell_\infty$-bound adversaries generally results in similar trends regarding human-like behavior with respect to the shape-texture bias as $\ell_2$-bound adversarial training. However, $\ell_\infty$-robust models perform better on high-frequency, and worse on low-frequency data.
    \item Observations made by prior work on $\ell_2$-bound ResNet-50 scale to other CNNs and Transformers relative to parameter sizes.
    \item Although Transformers also experience a drop in OOD performance after adversarial training, they perform better in OOD generalization and are more human-like than robust CNNs, and even outperform humans on many benchmarks.
    \item From the analysis of the images frequency spectra, we provide a possible explanation of why adversarial training can lead to a decay of model accuracy on OOD data.
    \item We also provide a possible explanation of why adversarial training reduces texture bias and increases shape bias.
\end{itemize}

\section{Related Work}
\label{sec:related_work}
This work focuses on the intersection between adversarial robustness and ``human-like'' behavior which we briefly sketch in this section.
\paragraph{Adversarial robustness.}
Neural networks have a tendency to overfit the training data distribution, which makes them fail to generalize beyond it. As a result, their predictions are often highly sensitive to small changes in input \cite{biggio2013,SzegedyZSBEGF13}, even if those changes are imperceptible and meaningless to humans. This phenomenon can be formally described as an adversarial attack, where the goal is to find an additive perturbation to the input sample that maximizes the loss function \cite{goodfellow2015explaining,madry2018towards,croce2020reliable,croce2020minimally}. To constraint attacks, perturbations are only sought within a specified radius $\epsilon$ (budget) of the original input. The radius is typically bounded by the $\ell_2$ or $\ell_\infty$-norm.

Adversarial attacks can be found in both white-box \cite{madry2018towards,croce2020minimally,croce2020reliable} and black-box \cite{liu2017delving,ilyas2018blackbox,Bhagoji2018Oct,andriushchenko2020} settings, with gradient-based attacks being particularly effective. Models that are not trained with adversarial defenses are typically only robust to low budgets attacks, if at all. Adversarial training (AT) \cite{madry2018towards} is a solution to this problem, as it trains the model on worst-case perturbations found during training, effectively making out-of-domain attacks become in-domain samples. However, this approach can result in overfitting to attacks used during training. Early stopping \cite{Zhang2020AttacksWD} and the addition of external (synthetic) data \cite{rebuffi2021data,gowal2021improving,wang2023better} have been proposed as effective solutions to address this problem.

However, adversarial robustness does not necessarily correlate with improved generalization and can even hurt it \cite{SaikiaSB21}. Supposedly again due to overfitting of training data, \eg models can still be susceptible to adverse weather conditions, image artifacts due to image compression, changes in lighting, etc.~\cite{hendrycks2019,lopes2020improving}.
\begin{figure}
    \centering
    \resizebox{\columnwidth}{!}{
         \begin{tabular}{@{}ccccc@{}}
         \textbf{colour (B\&W)} & \textbf{contrast} &\textbf{eidolon I} & \textbf{eidolon II} & \textbf{eidolon III}\\
         \includegraphics[width=0.32\columnwidth]{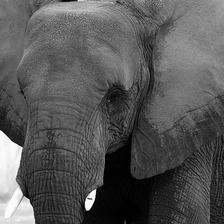} & 
         \includegraphics[width=0.32\columnwidth]{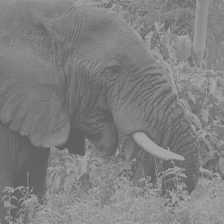} & 
         \includegraphics[width=0.32\columnwidth]{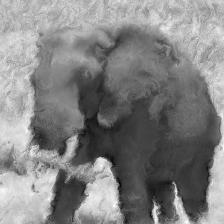} &
         \includegraphics[width=0.32\columnwidth]{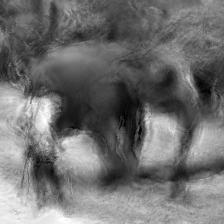} & 
         \includegraphics[width=0.32\columnwidth]{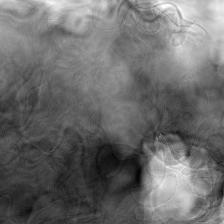} \\
         \textbf{false-colour} & \textbf{high-pass} & \textbf{low-pass} & \textbf{phase-scrambling} & \textbf{power-equalisation} \\
         \includegraphics[width=0.32\columnwidth]{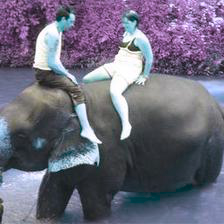} &
         \includegraphics[width=0.32\columnwidth]{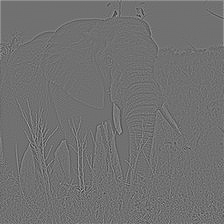} & 
         \includegraphics[width=0.32\columnwidth]{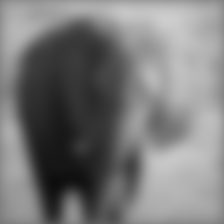} &
         \includegraphics[width=0.32\columnwidth]{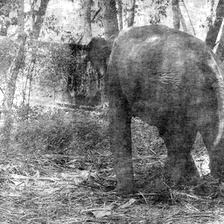} & 
         \includegraphics[width=0.32\columnwidth]{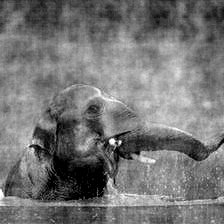} \\
         \textbf{rotation} & \textbf{uniform-noise} & & & \\
         \includegraphics[width=0.32\columnwidth]{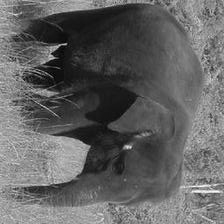} &
         \includegraphics[width=0.32\columnwidth]{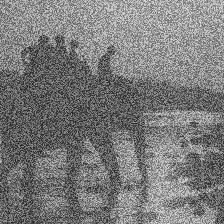} & & &\\
         \textbf{edge} & \textbf{silhouette} & \textbf{sketch} & \textbf{stylized}  & \textbf{cue-conflict (dog)} \\
         \includegraphics[width=0.32\columnwidth]{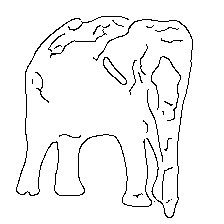} & 
         \includegraphics[width=0.32\columnwidth]{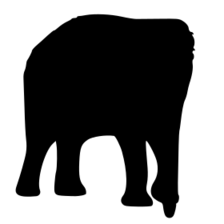} &
         \includegraphics[width=0.32\columnwidth]{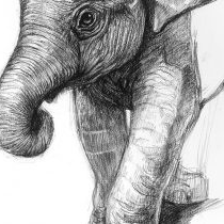} & 
         \includegraphics[width=0.32\columnwidth]{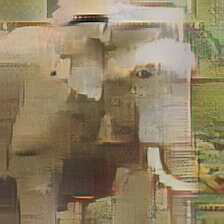} &
         \includegraphics[width=0.32\columnwidth]{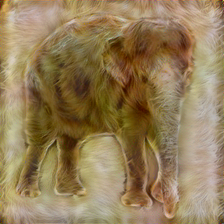} 
         \end{tabular}
     }
     \caption{OOD examples from \cite{geirhos2018imagenettrained,Geirhos2018Generalisation,Wang2019Learning,Geirhos2021_modelsvshumasn} for the ImageNet class ``elephant''.}
    \label{fig:examples}
\end{figure}
\paragraph{Measuring ``human-like'' behavior.}

\textit{Geirhos}~\etal propose to measure ``human-like'' reasoning via out-of-distribution (OOD) generalization to datasets and consistency in predictions with humans \cite{Geirhos2021_modelsvshumasn}.

Regarding OOD, they propose to benchmark against a set of 12 ImageNet modification datasets \cite{Geirhos2018Generalisation} at various intensities/conditions. At first glance, this may sound familiar to ImageNet-C \cite{hendrycks2019}, but benchmarks a different set of modifications: \textit{(the absence of) colour}, \textit{contrast (changes)}, \textit{eidolon~I/II/III}, \textit{false-colour}, \textit{high/low-pass (frequency filtering)}, \textit{phase-scrambling}, \textit{power-equalisation}, \textit{rotation}, \textit{uniform-noise}. Additionally, they propose to benchmark against a set of five OOD datasets aiming to identify the shape-texture-bias \cite{geirhos2018imagenettrained}: \textit{stylized}, \textit{edge}, \textit{silhouette}, \textit{texture/shape cue-conflict}, and \textit{sketch} (the latter provided by \cite{Wang2019Learning}). All datasets contain samples that belong to 16 ImageNet classes and are therefore classifiable by ImageNet models. For all datasets, the authors include a baseline obtained in lab settings over 4-10 human annotators. The \textit{cue-conflict} dataset is of particular interest, as neural networks are not only prone to overfit but - at least in the vision domain - they also tend to compute predictions based on details such as the texture of images rather than shapes, which does not align with human vision \cite{geirhos2018imagenettrained}. For example, an image of an elephant with an overlaid lion texture will most likely result in a prediction as ``lion'', while most humans would predict ``elephant'' as the true label when given the choice between both. It is worth noting that the authors also mention that ImageNet can be largely accurately classified solely based on texture. As such, ImageNet performance is insufficient as an indicator of ``human-like'' decision making, and \textit{Geirhos}~\etal propose to additionally report the \textit{cue-conflict} score to quantify this phenomenon. Examples of all datasets are shown in \cref{fig:examples}.

As an additional metric to accuracy, \cite{Geirhos2020error} propose to evaluate the agreement in predictions. In particular, they analyze false predictions (\textit{error consistency}) as well as the intersection rate of predictions where both humans and models have made a correct prediction (\textit{observed consistency}). 

The authors maintain a leaderboard of the most ``human-like'' models, which is currently dominated by Transformers such as \textit{ViT} \cite{dosovitskiy2021an,Zhai_2022_CVPR} and \textit{CLIP} \cite{pmlr-v139-radford21a}, or large convolutional neural networks \cite{yalniz2019billionscale, Kolesnikov2020Oct} - all being pre-trained on massive datasets. 
\section{Method}
\label{sec:method}
\begin{figure}
    \centering
    \includegraphics[width=\columnwidth]{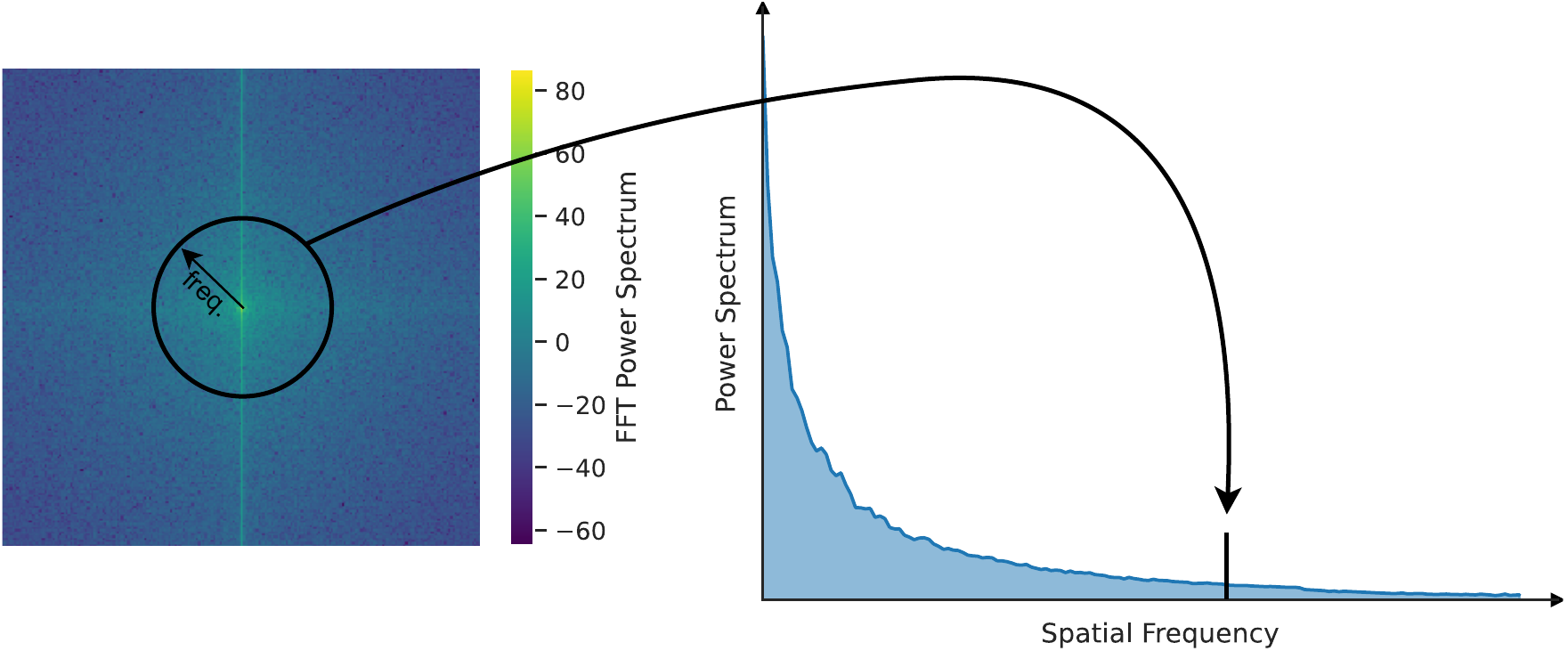}
    \caption{Visualization of how we obtain the spectrum plots. Each frequency measurement in the spectrum plot corresponds to the integral over the FFT power spectrum (frequency increases from the center to outer edges) up to that particular frequency. }
    \label{fig:fft_to_azimuth}
\end{figure}
To study the likeliness to human-like behavior of adversarially-trained models in greater detail, we use publicly available checkpoints and perform an analysis according to the setting proposed in 
\cite{Geirhos2021_modelsvshumasn}. We analyze pre-trained \textit{ResNet-18, ResNet-50, WideResNet-50-2} models trained against $\ell_\infty$-bound adversaries with $\epsilon\in\{0.5/255,1/255, 2/255,4/255,8/255\}$, and against $\ell_2$-bound adversaries with $\epsilon\in\{0.01,0.03,0.05,0.1,0.25,0.5,1,3,5\}$, and clean baselines (all provided by \cite{NEURIPS2020_24357dd0}). Further, we analyze \textit{XCiT-S/M/L12} Transformer models trained against $\ell_\infty$-bound adversaries with $\epsilon=4/255$ provided by \cite{debenedetti2023light} and a clean \textit{XCiT-S}\footnote{Clean pre-trained \textit{XCiT-M/L12} with the same configuration were not available.} baseline provided by \cite{rw2019timm}. Lastly, to better understand the differences between CNNs and Transformers, we also analyze a clean \textit{ConvMixer-768-32} \cite{Trockman2022} checkpoint, again obtained from \cite{rw2019timm}.
All models were trained on ImageNet \cite{imagenet} without any additional pre-training.

For all models, we measure the accuracy of all datasets by reporting the mean overall conditions in the dataset where average human performance was above 20\% accuracy. Lastly, we determine the \textit{observed} and \textit{error consistency} against human annotators again as a mean over all datasets and conditions. As there are multiple annotators per dataset, we calculate consistencies against each annotator and report the mean.

We first provide a more extensive evaluation of models that have been trained using $\ell_2$-AT. Then, we provide insights on how models trained with $\ell_2$-AT behave compared to models that are trained using $\ell_\infty$-AT. Comparing these two training types is not straightforward, due to the different types of perturbations they cause. As the $\ell_2$-norm penalizes the euclidean distance, perturbations can locally be more severe than under $\ell_\infty$. Yet, if the perturbation magnitude increases the area of perturbations has to decrease under $\ell_2$-norm, while attacks under the $\ell_\infty$-norm can add perturbations to the entire image without constraints except for the magnitude. Thus, there are multiple options for choosing comparable budgets between the two norms. We choose a straightforward way and select budgets for both norms that approximately result in the same clean accuracy shown in \cref{tab:eps_cmp}. As we have more checkpoints for $\ell_2$-AT training, we only use a subset of those in the following analysis.
%
\begin{table}
    \centering
    \scriptsize
    \caption{Our chosen $\epsilon$ budgets for comparisons between $\ell_2$- and $\ell_\infty$-bound training.}
    \label{tab:eps_cmp}
    \begin{tabular}{c|cccc}
    \toprule
    $\ell_2$ & $0.1$&$1$ &$3$ &$5$\\
    \midrule
    $\ell_\infty$ & $0.5/255$&$1/255$&  $4/255$ &  $8/255$\\     
    \bottomrule
    \end{tabular}
\end{table}

Next, we compare the behavior of CNN and Transformer models under these training settings. Based on these experiments, we then discuss whether AT is an effective tool to induce a more human-like behavior in trained models. Finally, we impose a frequency perspective on OOD performance and shape bias under AT. To back this analysis, we plot the frequency distribution for each OOD dataset, and clean ImageNet validation samples belonging to the same classes. Then we compare each OOD distribution to the clean distribution to understand where shifts in the frequency distribution are located. We obtain the frequency distribution plots as introduced in \cite{Durall_2020_CVPR}: we compute the log-scaled FFT power spectrum and compute the radial integral under increasing frequency resulting in a frequency power distribution (\cref{fig:fft_to_azimuth}). For comparability, we scale the resulting distributions by their integral.
\begin{figure}
    \centering
    \includegraphics[width=\linewidth]{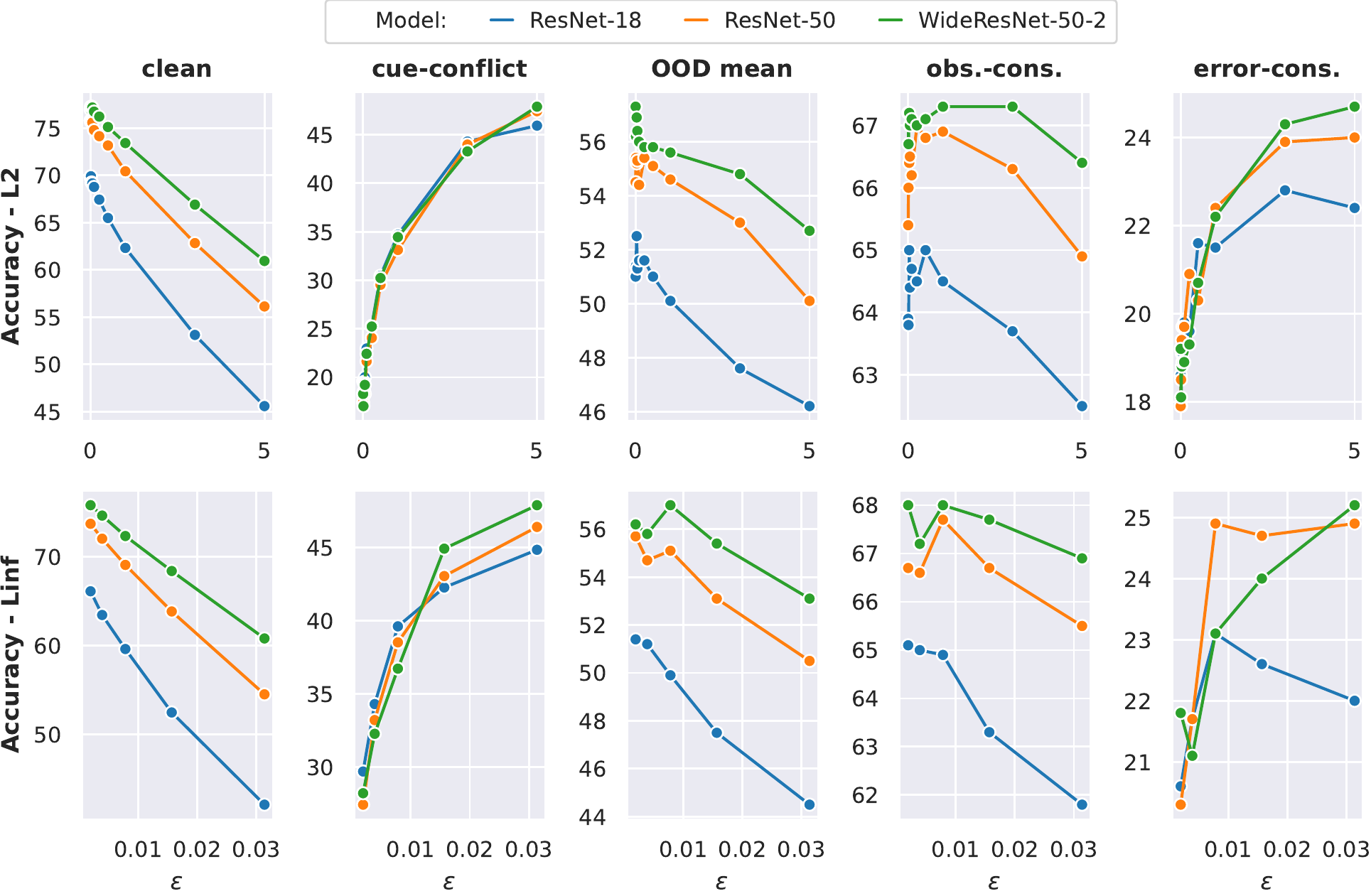}
    \caption{Performance of $\ell_2$ vs. $\ell_\infty$-AT-trained \textit{ResNet-18, ResNet-50, WideResNet-50-2} on clean data, texture/shape bias cue-conflict datasets, the average mean of all OOD datasets, and observed/error consistency compared to humans under increased training attack budget $\epsilon$.}
    \label{fig:budget_trend}
\end{figure}
\begin{figure*}
    \centering
    \includegraphics[width=\linewidth]{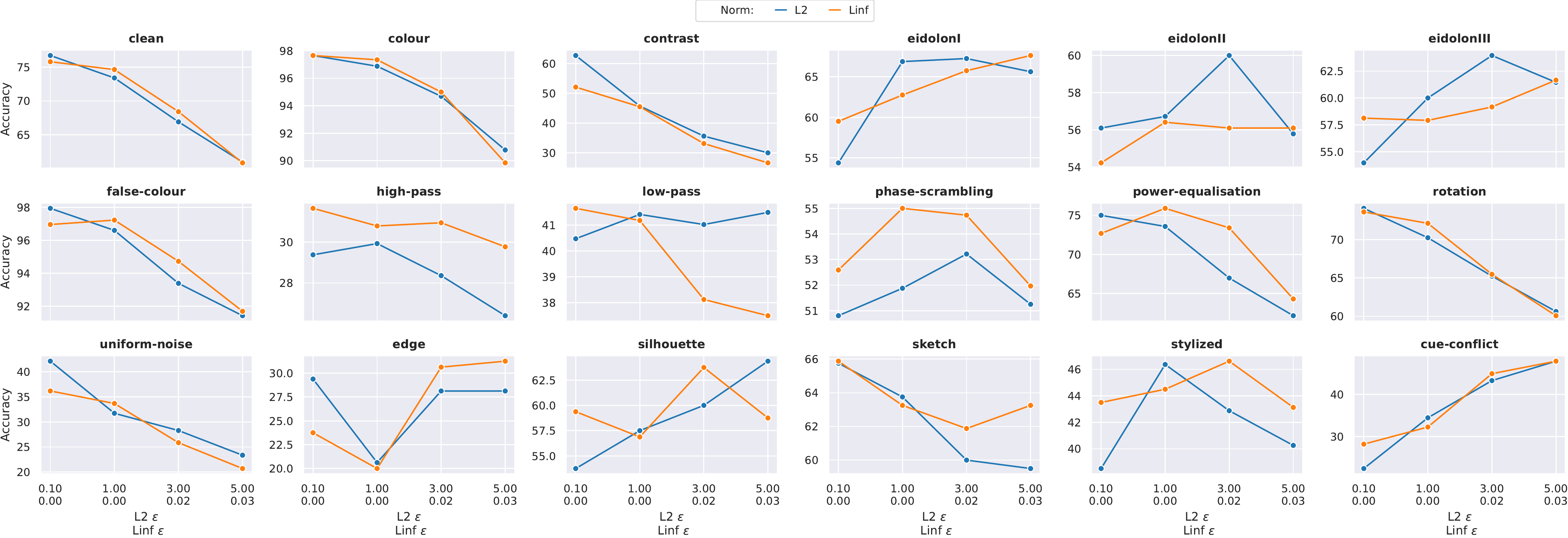}
    \caption{Comparison of performance on OOD datasets between robust \textit{WideResNet-50-2} trained against $\ell_2$- (upper $\epsilon$ values) and $\ell_\infty$-bound (lower $\epsilon$ values) adversaries under increasing training budget $\epsilon$. $\epsilon$ are selected in a way that clean accuracy approximately matches between the norms.}
    \label{fig:linf_vs_l2}
\end{figure*}
%
\section{Results}
\label{sec:results}
\paragraph{Width and depth of $\ell_2$ CNNs.}
First, we want to evaluate the effect of $\ell_2$-training on CNN architectures: \textit{Wide-ResNet-50-2}, \textit{ResNet-50} and \textit{ResNet-18} (\cref{fig:budget_trend}, top row), which allows investigating the effect of $\ell_2$-AT depending on model depth and width. 

We observe that increasing both, width and depth improves clean performance, and performance on all OOD datasets, as well as the observed and error consistency. As such, \textit{Wide-ResNet-50-2} performed best on clean performance, OOD mean, and observed/error consistency. It is also worth noting, that switching from \textit{ResNet-50} to \textit{Wide-ResNet-50-2} has a smaller impact on performance than switching from \textit{ResNet-18} to \textit{ResNet-50}. Also, we observe that in some cases \textit{ResNet-18} shows opposite trends with respect to training budget than 50-layer deep \textit{ResNets}, \eg for \textit{power-equalisation}. Still, \textit{ResNet-18} performs best on the \textit{edge} dataset for large training budgets (not shown due to space limitations). Overall, 
this suggests, that increasing parameterization of $\ell_2$-bound adversarially-trained models correlates with an increase in human-like behavior.

\paragraph{$\ell_2$- vs. $\ell_\infty$-bound Adversarial Training.}
Next, we compare how $\ell_2$-AT relates to $\ell_\infty$-AT with respect to human-like reasoning. 
Exemplarily, we analyze the trend under comparable budgets of a \textit{WideResNet-50-2} (\cref{fig:linf_vs_l2}). We observe that on some datasets there is barely any perceivable difference as the budget increases (\textit{colour, contrast, false-colour, uniform-noise, rotation, cue-conflict}), but there are cases where one norm or the other clearly performs better. $\ell_\infty$-robust models seem to be more robust against \textit{high-pass}, \textit{phase-scrambling}, and \textit{power-equalisation}. On the other hand, $\ell_2$-robust models appear to perform better on \textit{low-pass}, and \textit{eidelonII/III}. Lastly, there are also some inconclusive settings where one or the other performs better depending on the budget (\textit{silhouette, eidolon~I, stylized}). Besides \textit{cue-conflict}, none of the OOD categories clearly benefit from AT for WideResNet-50-2. These observations only partly transfer to other CNN architectures in \cref{tab:my_label}. In general, \cref{fig:budget_trend} (bottom) shows a similar trend for $\ell_2$ and $\ell_\infty$-AT, and all results support the finding that the \textit{cue-conflict} score increases consistently under both types of AT, \ie the behavior becomes more human-like towards shape bias in both cases. Therefore, we conclude that the more commonly used $\ell_\infty$-AT is equally effective in inducing human-like behavior in CNNs, with respect to \textit{cue-conflict}, and \textit{consistency}. 
\begin{table}
    \centering
    \scriptsize
    \caption{Comparison between parameters of analyzed models.}
    \label{tab:params}
    \begin{tabular}{llr}
    \toprule
    \textbf{Model} & \textbf{Inductive Bias} & \textbf{Parameters}\\
    \midrule
         ResNet-18 & CNN & $10.4$ M\\
         ResNet-50 & CNN & $25.6$ M\\
         WideResNet-50-2 & CNN & $68.9$ M \\
         \midrule
         ConvMixer-768-32 & Hybrid & $21.2$ M\\
         \midrule
         XCiT-S12 & Transformer & $26.3$ M\\
         XCiT-M12 & Transformer & $46.4$ M\\
         XCiT-L12 & Transformer & $103.8$ M\\
    \bottomrule
    \end{tabular}
\end{table}
\paragraph{CNNs vs. Transformers.}
Finally, we expand our analysis to Transformer architectures (\textit{XCiT}) for which we only report clean and $\ell_\infty$-training performance. 
\begin{table*}
    \centering
    \small
    \caption{Results in [\%] of generalization performance and consistency with human predictions/errors. For robust models we only report $\ell_2, \epsilon=3$ ($\ell_2$) and $\ell_\infty, \epsilon=4/255$ ($\ell_\infty$) for brevity. Models without adversarial training are highlighted in gray. \textbf{Bold} values indicate the best performance amongst all models.}
    \label{tab:my_label}
\resizebox{\linewidth}{!}{

\begin{tabular}{l|c|cccccccccccc|ccccc|c|cc} 
\toprule
 & & \multicolumn{18}{c|}{\textbf{Out-of-distribution Performance}} & \multicolumn{2}{c}{\textbf{Consistency}}\\
\textbf{Model}   & \textbf{Clean} & colour    & contrast  & eidolon       & eidolon      & eidolon     & false   & high      & low       & phase & power & rotat.       & uniform  & edge           & silh.     & sketch         & styliz.       & cue   & \textbf{Mean}            & correct & error  \\
        &       &           &           &     I   &   II    &   III   & colour   & pass      & pass       & scr. & equal. &        & noise  &            &      &          &        & conflict   &             &  &  \\
\midrule
\rowcolor{gray!30}R18              & 69.79          & 95.47          & 71.88          & 47.50          & 51.88          & 49.38          & 93.39          & 32.66          & 37.73          & 48.21            & 61.25              & 68.36          & 
34.22          & 18.12          & 41.88          & 59.00          & 36.00          & 19.61          & 51.00          & 63.90                & 18.60              \\
R18 ($\ell_2$)          & 53.12 & 86.25 & 27.50 & 60.25 & 49.53 & 51.46 & 85.27 & 24.14 & 35.39 & 47.50 & 51.96 & 55.23 & 20.78 & 27.50 & 61.25 & 51.12 & 39.50 & 44.30 & 47.60 & 63.70 & 22.80 \\
R18 ($\ell_\infty$)             & 52.49          & 84.69          & 23.62          & 61.12          & 50.94          & 51.67          & 83.57          & 25.86          & 35.55          & 47.05            & 56.07              & 55.23          & 18.83          & 26.88          & 56.88          & 50.88          & 40.62          & 42.27          & 47.50          & 63.30                & 22.60              \\
\midrule

\rowcolor{gray!30}R50              & 75.80          & 97.19          & 83.62          & 49.12          & 52.66          & 51.04          & 95.62          & 33.67          & 38.98          & 49.11            & 70.71              & 73.91          & 37.97          & 23.75          & 48.12          & 61.25          & 34.38          & 17.42          & 54.50          & 65.40                & 17.90              \\
R50 ($\ell_2$)          & 62.83 & 92.81 & 32.12 & 66.12 & 56.41 & 62.71 & 90.71 & 26.17 & 40.31 & 53.84 & 63.57 & 63.75 & 26.09 & 25.62 & 60.62 & 59.38 & 41.75 & 43.98 & 53.00 & 66.30 & 23.90 \\
R50 ($\ell_\infty$)            & 63.86          & 91.25          & 29.25          & 64.25          & 54.37          & 57.50          & 91.07          & 30.70          & 38.52          & 53.39            & 68.04              & 64.06          & 26.25          & 25.62          & 58.75          & 60.50          & 43.25          & 43.05          & 53.10          & 66.70                & 24.70              \\
\midrule
\rowcolor{gray!30}WRN50-2     & 76.97          & 98.28          & 82.38          & 51.00          & 54.69          & 54.17          & 97.23          & 34.92          & 40.62          & 50.98            & 75.18              & 75.39          & 42.27          & 28.75          & 56.88          & 64.12          & 36.50          & 18.28          & 57.30          & 67.20                & 19.20              \\
WRN50-2 ($\ell_2$) & 66.90 & 94.69 & 35.62 & 67.25 & 60.00 & 63.96 & 93.39 & 28.36 & 41.02 & 53.21 & 66.96 & 65.23 & 28.28 & 28.12 & 60.00 & 60.00 & 42.88 & 43.28 & 54.80 & 67.30 & 24.30 \\
WRN50-2 ($\ell_\infty$)   & 68.41          & 95.00          & 33.12          & 65.75          & 56.09          & 59.17          & 94.73          & 30.94          & 38.12          & 54.73            & 73.39              & 65.47          & 25.86          & 30.63          & 63.75          & 61.88          & 46.62          & 44.92          & 55.40          & 67.70                & 24.00              \\
\midrule
\rowcolor{gray!30}ConvMixer-768-32 & 80.16 & \textbf{99.22} & 98.00 & 50.62 & 56.72 & 56.25 & 98.04 & 39.77 & 43.91 & 56.43 & 86.25 & 80.23 & \textbf{56.02} & 26.88 & 64.38 & 70.75 & 44.50 & 22.73 & 63.30 & 69.50 & 19.50 \\
\midrule
\rowcolor{gray!30}XCiT-S12      & \textbf{81.97} & {98.91} & \textbf{98.88} & 55.12          & 59.38          & 64.17          & \textbf{98.75} & \textbf{69.84} & \textbf{46.72} & \textbf{62.14}   & \textbf{91.07}     & \textbf{81.41} & {55.62} & \textbf{37.50} & 61.88          & 71.12          & \textbf{57.75} & 25.55          & \textbf{68.90} & 70.90                & 19.50              \\
XCiT-S12  ($\ell_\infty$)               & 72.34          & 96.88          & 47.62          & 66.50          & 58.91          & 61.04          & 96.88          & 36.95          & 39.77          & 56.61            & 82.14              & 70.70          & 40.47          & 31.87          & 63.75          & 70.75          & 48.75          & 46.80          & 60.60          & 70.00                & 24.10              \\
XCiT-M12 ($\ell_\infty$)                & 74.04          & 97.34          & 48.25          & 66.88          & 60.16          & 62.29          & 96.96          & 36.80          & 39.06          & 57.59            & 81.43              & 70.86          & 41.17          & 26.25          & 66.88          & 71.00          & 52.62          & 47.27          & 60.90          & 70.40                & \textbf{24.80}     \\

 XCiT-L12 ($\ell_\infty$)                & 73.76          & 98.12          & 47.38          & \textbf{69.38} & \textbf{60.62} & \textbf{64.58} & 98.66          & 41.95          & 41.72          & 58.21            & 84.11              & 70.62          & 42.27          & 35.62          & \textbf{69.38} & \textbf{74.00} & 54.12          & \textbf{48.83} & 63.40          & \textbf{71.10}       & 22.70              \\

\midrule
\rowcolor{blue!10}\textbf{Humans}                          & -            & 88.67          & 66.09          & 60.75          & 58.28          & 63.91          & 88.82          & 46.43          & 56.09          & 55.11            & 75.89              & \textbf{84.51}          & 55.37          & \textbf{87.12}          & \textbf{75.31}          & \textbf{91.62}          & 47.12          & \textbf{77.55}          & -              & -                    & -                  \\
\bottomrule
\end{tabular}
}
\end{table*}%
On clean training, even the smallest Transformer (\textit{XCiT-S12}) which has a comparable number of parameters to \textit{ResNet-50} (\cref{tab:params}), performs significantly better than the largest CNN. Contrary to CNNs it is also able to surpass human performance on \textit{eidolon II/III}, \textit{high-pass} (with an impressive improvement of approx.~35\% above CNNs), \textit{phase-scrambling, power-equalisation, uniform-noise}, and \textit{stylized}.
Under AT, we largely see the same shift as with CNNs with one exception: while AT improves \textit{stylized} performance of CNNs, it decreased it on Transformers. Still, Transformers achieve higher accuracies than humans in this category.
Of all studied models, the adversarially-trained \textit{XCiT-L12} performs best on \textit{eidolon I-III}, \textit{silhouette, sketch, and cue-conflict}. However, it is also worth noting that it contains 50\% more parameters compared to the largest CNN we analyze. In general, we can not conclude that more parameters are always better as we \eg see some reduction in error consistency from robust \textit{XCiT-S/M12} to \textit{XCiT-L12}. Further, the clean \textit{ConvMixer} which contains no self-attention but patch-embeddings, shows also an increased \textit{cue-conflict}. Generally, there is a trade-off between CNNs and Transformers in almost all studied datasets. We hypothesize that patch-embeddings may naturally be slightly shifted toward shape bias compared to CNNs.

\paragraph{Is adversarial training a good option to achieve human-like reasoning?}
While AT does improve \textit{cue-conflict} significantly and shifts the internal decision process toward human-like shape bias behavior, it also decreases OOD  performance across many datasets. Most notably, AT causes a significant drop in robustness to changes in \textit{contrast}, \textit{rotation}, and \textit{uniform noise} compared to clean training. 
Interestingly, it also always reduces \textit{high-pass} performance. In the case of \textit{XCiT}-models this performance is slightly worse than for humans after AT, although the clean model significantly outperformed humans (by approx.~23\%). 
We see the largest OOD drops in \textit{XCiT}, while the \textit{ResNets} show only minor impairments. Based on these findings, AT alone is not sufficient to shift models toward human-like reasoning in all aspects. In the next section, we investigate the frequency spectra of OOD samples and show that they provide an indication of whether AT can, in principle, help to increase performance.
\begin{figure*}
    \centering
    \includegraphics[width=\linewidth]{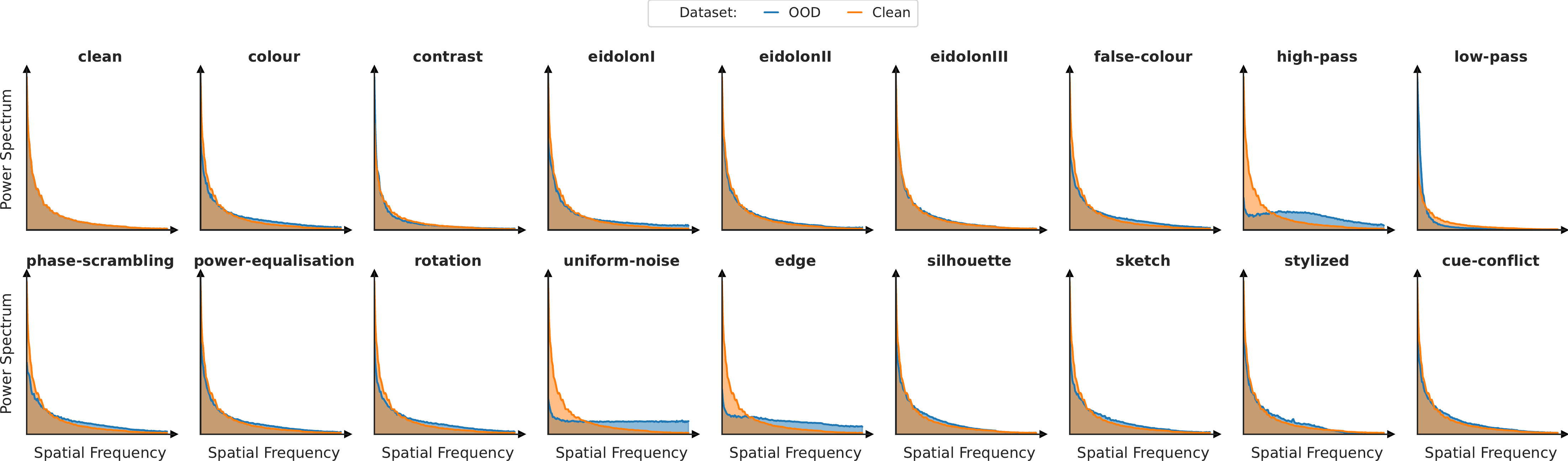}
    \caption{Frequency distribution of the utilized OOD datasets in comparison to comparable ImageNet validation samples (clean). Distributions are normalized by their integral. Frequency increases along the X-axis.}
    \label{fig:ood_spectrum}
\end{figure*}
\section{A Frequency Perspective on Adversarial Training and Out-Of Distribution Data}
In \cref{fig:ood_spectrum}, we plot for all considered OOD image categories their frequency power spectra, radially integrated as described in \cref{fig:fft_to_azimuth}, and compare the frequency spectra to the spectrum of the clean training images. From this comparison, it is apparent that some OOD categories deviate heavily from the natural image distribution in terms of their spectra. This is obviously true for \textit{high-pass} and \textit{low-pass} images as well as for \textit{uniform-noise} and \textit{edge}, where the differences are particularly strong in the high-frequency regime, but it is also visible for \textit{contrast}, \textit{rotation}, and \textit{power equalization}, or \textit{phase scrambling}, with significant differences in the lowest frequencies. Although adversarial attacks might slightly alter the frequency spectrum of attacked images, they are $\epsilon$ bounded and will therefore not significantly change the frequency distribution over all samples. Thus, it would be natural that AT (\ie adding more training samples with a spectral distribution similar to the one of clean images) would harm the transferability of models to such out-of-domain categories. In fact, \cref{tab:my_label} shows exactly this trend: both types of AT cause a consistent decay in classification accuracy for the OOD categories  \textit{high-pass}, \textit{low-pass}, \textit{uniform-noise}, \textit{contrast}, \textit{rotation} and \textit{power equalization}. When the differences are in the low-frequency regime as for \textit{contrast}, the decay seems to be particularly strong. This observation supports the findings by \cite{SaikiaSB21} that AT can harm robustness to other corruption types and provides an initial explanation: Adding more training samples from the original spectral distribution can harm the generalization to other diverging spectral distributions. 

\cref{fig:ood_spectrum} also shows that some OOD categories have power spectra that are quite similar to the spectra of the original data (and thus of adversarial examples). For these categories, \eg \textit{eidolon I}, \textit{eidolon II}, \textit{eidolon II}, \textit{false colour} or \textit{cue-conflict}, AT does not lead to a decay in accuracy but can even lead to improvement in some cases. In the following, we will discuss in which cases we might expect this improvement.  

From the above observation, we see that the OOD data should share some important properties with the clean data to benefit from AT, \ie the frequency distribution should not differ too much. At the same time, it has been argued that convolutional neural networks tend to decide based on texture information \cite{geirhos2018imagenettrained}, which is local and rather mid to high-frequency. Thus, adversarial examples can attack such models by slightly altering the image in these frequency bands. While this may vary by dataset \cite{Maiya2021,AbelloHW21,BernhardMMBCSR21}, at least some high-frequency is always present as  \eg adversarial attacks can be detected in the frequency spectrum~\cite{Harder2021}.

To compensate for these attacks, robust models desensitize to high-frequency and instead shift their decisions towards global cues that involve low-frequency information, which can typically be observed in FFT-spectra of perturbations after AT (\eg \cite{Grabinski2022Nov}, Fig.~8). The desensitization of high-frequencies during training also results in more robust models, as shown from various perspectives such as injecting noise patches to inputs \cite{lopes2020improving}, blurring feature-maps \cite{NEURIPS2020_f6a673f0}, splitting and regularizing frequency information \cite{SaikiaSB21}, or low-pass filtering intermediate feature-maps \cite{Grabinskilowcut22} during training.
There seem to be sufficient indicators to reasonably assume that shifting the decisions toward low-frequency information by removing the focus from high-frequency is at least a necessary ingredient of robustness. Clearly, AT encourages this shift, which can also be seen in weights of convolution filters of robust models \cite{Gavrikov_2022_CVPR,Gavrikov_2022_CVPRW}.

Likewise, texture bias can also be analyzed from a frequency perspective. Textures contain high-frequency information while shapes can not be represented without low-frequency bands. As non-robust neural networks naturally prefer high-frequency information for predictions they reason based on textures. Under AT, models rely less on local high-frequency information and prioritize the lower-frequent information, that encodes global structures such as shapes. This effect can be well seen in the \textit{cue-conflict} performance where images contain both types of information in the images, and models can choose which information to prioritize. From an information perspective alone, both choices would acceptable.

Ultimately, this perspective does not explain all findings \underline{and other mechanics may influence the decision process}. For example, \textit{stylized} performance improves under AT for CNNs while the accuracy of Transformers, starting at a higher level, is decreasing. It can just provide an intuition of why the model decisions learn to shift towards a more global, shape bias  - given that the overall spectral distribution remains very similar to the original training data distribution in the \textit{cue-conflict} category. 
\section{Conclusion}
\label{sec:conclusion}

We have extended previous experiments that studied the influence of $\ell_2$-AT on the reasoning of neural networks in comparison to human reasoning. Our findings indicate, that previous observations scale to $\ell_\infty$-AT, other CNNs, and even Transformers. In general, we find that robust Transformers appear to be more similar to human reasoning than CNNs as they perform better on OOD datasets and increasingly reason based on shape information. Still, AT results in degradation against some corruptions that do not seem to affect humans or models trained without AT.  Finally, we propose an explanation of why AT enforces shape bias from a frequency perspective: AT seems to hurt generalization against OOD datasets where the spectral distribution significantly diverges from the training data. In other cases, AT causes the model to shift its decision from local high-frequency information to global shape information, which better resembles the behavior of humans.

\clearpage
{\small
\bibliographystyle{ieeetr_fullname}
\bibliography{main}
}

\end{document}